\documentclass[runningheads]{llncs}
\usepackage[T1]{fontenc}
\usepackage{graphicx}
\usepackage{adjustbox}
\usepackage{hyperref}
\usepackage{amsmath}
\usepackage{algorithm}
\usepackage{algpseudocode}
\usepackage[thin, , thinc]{esdiff}
\usepackage{subcaption}
\usepackage{caption}
\usepackage{framed}
\usepackage{multirow}
\usepackage{tabularx}
\usepackage{indentfirst}

\usepackage{tikz}
\usetikzlibrary{arrows.meta}
\usepackage{array}

\begin{document}
\title{Adaptive Bias Generalized Rollout Policy Adaptation on the Flexible Job-Shop Scheduling Problem}
\titlerunning{ABGNRPA on FJSSP}
%

\author{Lotfi Kobrosly\inst{1,2} \and Marc-Emmanuel Coupvent des Graviers\inst{1} \and 
Christophe Guettier\inst{1} \and Tristan Cazenave \inst{2}}

\authorrunning{L. Kobrosly et al.}
%
\institute{Safran Electronics and Defense, France \and
LAMSADE, Université Paris Dauphine-PSL, Place du Maréchal de Lattre de Tassigny, Paris, France}
\maketitle              
\begin{abstract}
The Flexible Job-Shop Scheduling Problem (FJSSP) is an NP-hard combinatorial optimization problem, with several application domains, especially for manufacturing purposes. The objective is to efficiently schedule multiple operations on dissimilar machines. These operations are gathered into jobs, and operations pertaining to the same job need to be scheduled sequentially. Different methods have been previously tested to solve this problem, such as Constraint Solving, Tabu Search, Genetic Algorithms, or Monte Carlo Tree Search (MCTS). 
We propose a novel algorithm derived from the Generalized Nested Rollout Policy Adaptation, developed to solve the FJSSP. We report encouraging experimental results, as our algorithm performs better than other MCTS-based approaches, even if makespans obtained on large instances are still far from known upper bounds.

\keywords{Monte Carlo Tree Search \and Generalized Nested Rollout Policy Adaptation \and Policy Gradient Descent \and Markov Decision Process \and Flexible Job-Shop Scheduling Problem}

\end{abstract}
\section{Introduction}
\label{intro}

The Flexible Job-Shop Scheduling Problem (FJSSP), as an extension of the more widely known Job-Shop Scheduling Problem (JSSP), has been studied for years in the research community. This is due to its proximity to manufacturing, aeronautics, and medicine applications \cite{saqlain2023monte,dauzere2024flexible}, but also because of the computational challenge it poses, since it qualifies as an NP-hard problem \cite{pezzella2008genetic,xie2019review,watson2003problem,mastrolilli2011hardness}. The usual branch-and-bound methods that work for the JSSP do not generalize well to the FJSSP \cite{dauzere2024flexible}, suggesting the need for alternative methods to solve this problem.

%
The goal of the FJSSP is to assign ordered operations of different jobs sequentially on the available machines, while minimizing a certain objective function\cite{fera2013production}. 
%
To address the problem, Monte Carlo Tree Search (MCTS), which can be considered as a form of Reinforcement Learning \cite{vodopivec2017monte} that quantifies the tradeoff between exploration and exploitation through probing \cite{browne2012survey}, qualifies as a promising solving method. Indeed, we can consider the FJSSP as a sequential decision problem. MCTS methods proved successful in solving this type of problems, such as one-player and two-player games \cite{cazenave2021monte,silver2016mastering} or in Operational Research (OR) problems like the Vehicle Routing Problem \cite{cazenave2020monte}.
%

In this paper, we present a new approach called the \emph{Adaptive Bias Generalized Nested Rollout Policy Adaptation (ABGNRPA)} on the FJSSP and compare it to the Generalized Nested Rollout Policy Adaptation (GNRPA) from which it stems, as well as to other variations of the MCTS. \hyperref[background]{Section ~\ref*{background}} describes the FJSSP problem and the GNRPA while presenting some of the previous work on these subjects. \hyperref[abgnrpa]{Section \ref*{abgnrpa}} presents the ABGNRPA and finally \hyperref[results]{section~\ref*{results}} showcases our experimental procedure and results.

\section{Background and motivation}
\label{background}

\subsection{The Job-Shop Scheduling Problem}
\label{jssp}

\begin{figure}
    \centering
    \includegraphics[width=\linewidth]{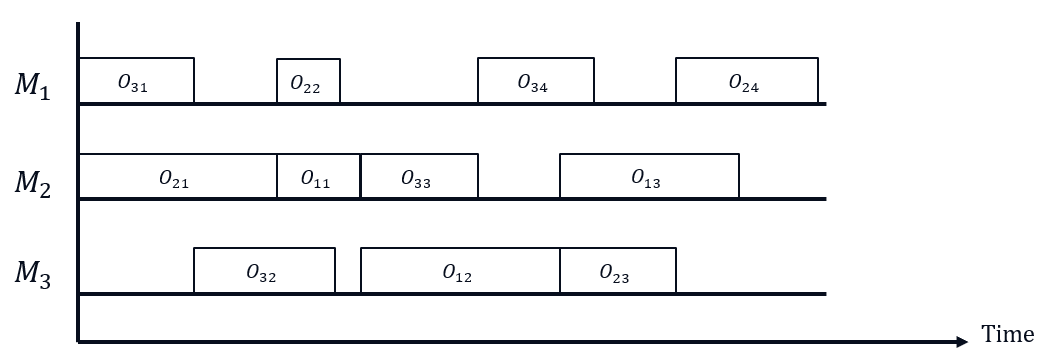}
    \caption{Gantt's diagram of an example of a JSSP instance. The operations denoted by $O_{ij}$ are assigned to machines $M_l$ while respecting the constraints of order of operations in a job, of compatibility with machines and no interruption of the processing of operations. Blank spaces in a machine's line of processing represent inactivity.}
    \label{fig:jssp}
\end{figure}

The JSSP is a combinatorial optimization and scheduling problem. It is a theoretical representation of several industrial use cases where we schedule the processing of several operations in an orderly fashion to optimize a cost function. The formalized setting of the JSSP can be defined as follows, with a visual representation available in \hyperref[fig:jssp]{figure \ref*{fig:jssp}}:

\begin{itemize}
    \item[-] A set of $n$ jobs $\mathcal{J} = \{ \mathcal{J}_i ~ | ~ i \in \{1..n\}\}$.
    \item[-] A set of $m$ machines, $\mathcal{M} = \{m_l, l \in \{1..m\}\}$.
    \item[-] For every job $\mathcal{J}_i, ~ i \in \{1..n\}$, there are $n_i$ operations $\mathcal{J}_i = \{o_{ij}, ~ j \in \{1..n_i\}\}$. 
    \item[-] The operations in each job are ordered, which means that for operations $o_{ij},o_{ik} \in \mathcal{J}_i, ~ j, k \in \{1..n_i\}$ and $j < k$, $o_{ij}$ must be processed before $o_{ik}$.
    \item[-]  Every operation $o_{ij}$ is defined in the problem's instance to be compatible with only one machine $m_l, ~ l \in \{1..m\}$ with a fixed processing time $d_{ij}$.
    \item[-] Once the processing of an operation $o_{ij}$ starts on a machine $m_l$, it is done until its completion, \emph{i.e} it cannot be interrupted.
    \item[-] The objective is to optimize a cost function.
\end{itemize}

Several approaches have been adopted to address JSSP. Some of them model it as a Mixed Integer Linear Program \cite{meng2020mixed}, while others considered solving it with Constraint Programming (CP) \cite{zhou1996constraint,da2022industrial}, Tabu Search \cite{nowicki2005advanced} or other local search techniques \cite{beck2011combining}. More recently, Deep Reinforcement Learning (DRL) has been used as one of the most prominent approaches considering the JSSP as a sequential decision problem \cite{liu2020actor,zhou2020deep}. Graph Neural Networks have gained a good amount of consideration, whether used individually or combined with DRL \cite{hameed2020reinforcement} or a randomized $\epsilon$-greedy that can select the second best action \cite{abgaryan2024randomized}.

\subsection{The Flexible Job-Shop Scheduling Problem}
\label{fjssp}

The FJSSP is an extension of the JSSP, where in addition to the aforementioned setting, the assignment of operations on machines becomes an extra decision. Indeed, each operation can be processed by more than one machine instead of being statically allocated on a unique one as described in \hyperref[jssp]{section \ref*{jssp}}. Every couple of operation and machine $(o_{ij}, m_l)$ is associated with a processing time $d_{ijl}$ if they are compatible. These additions enlarge the search space as the choice of the assignment of an operation on a machine changes the duration, remove homogeneity within the problem structure, for instance where symmetry breaking is harder to apply, and therefore make the problem more difficult to approach. Solving it requires assigning operations to machines, on top of ordering them and setting their start times. The objective can vary from the most studied criterion of the minimization of the makespan \emph{i.e}, the total execution time \cite{dauzere2024flexible} to other ones \emph{e.g} the minimization of the workload at all instants $t$ \cite{chaudhry2016research} or the completion times of jobs \cite{de2010improved}. In our case, we focus on makespan minimization, which is more documented, providing us with more benchmarks to compare with our results.

%
The FJSSP can be represented with different approaches in addition to some of those applicable to the JSSP \cite{dauzere2024flexible,kasapidis2021flexible,naderi2022critical}. Several solving methods have been developed ranging from using Representation Learning via Graph Neural Networks \cite{park2021learning,song2022flexible} to Deep Reinforcement Learning \cite{zhang2020learning}, Memetic Algorithm \cite{luo2020efficient} and Genetic Algorithm \cite{de2010improved}, considering the problem as a Markov Decision Process (MDP) \cite{bellman1957markovian,park2021learning}. For the Dynamic FJSSP, another variant of the FJSSP, a time-based approach is adopted as operations and jobs can be added during the processing of already existing operations \cite{zhou2020deep,zhang2020evolving}.

\subsection{Generalized Nested Rollout Policy Adaptation}
\label{gnrpa}

Stochastic methods are useful for searching for good solutions, typically on an NP-hard problem such as the FJSSP. We can consider the FJSSP as a sequential decision making problem modeled by a search tree (see \hyperref[model]{section \ref*{model}} for the modeling details). Thus, a MCTS approach is a good candidate, in particular, the Generalized Nested Rollout Policy Adaptation (GNRPA) \cite{cazenave2021generalized}. In the search tree, a node represents a state of the advancement of a solution, and an arc from a parent node to a child represents an action.

In the general setting of a GNRPA, a rollout, or a playout, is one path through the search tree, from its root to a leaf. The choice of actions to generate this path is operated using a policy \cite{li2017deep} that assigns a weight to every action independently from the state. The weight defines the probability of selecting the corresponding action from available ones in the current state, using a \emph{softmax with temperature} function. The state-independent feature means that even if the set of possible actions differs from one state to another, an action has the same weight in all action sets in which it appears. Changes that occur to this weight are propagated to other action sets. The informal description of a \emph{level 1} GNRPA is as follows \cite{cazenave2021generalized}:

\begin{enumerate}
    \item The policy is initialized with the same value for all actions.
    \item A rollout is conducted through the search tree using the policy as a weighted random choice on the actions' set to select its next move. For every action, we compute a \emph{bias} $\beta_{action}$ statically via a \emph{heuristic} \cite{sentuc2021generalized}. For the FJSSP, we would refer to them as $\beta_{ijl}$ for job $i$, operation $j$ and machine $l$. This bias helps to initialize the probabilities of the actions.
    \item At every state, we compute the probabilities of the possible actions, according to the formula in \hyperref[weights_gnrpa]{equation~\ref*{weights_gnrpa}}, and the weighted randomized choice is performed on a softmax computation with $\tau$ a \emph{temperature}.
    \item At the end of this playout, we compare to the last best sequence and take the best out of the two based on their respective scores. Through this sequence, the weights of the encountered possible actions are updated, as shown in \hyperref[weight_update_gnrpa]{equation~\ref*{weight_update_gnrpa}}, where $\alpha$ is a learning rate. $\delta_{aa'} = 1$ for the chosen action in the trajectory $a=a'$ (assigning operation $j$ of job $i$ on machine $l$), thus increases its weight, and $\delta_{aa'} =0$ otherwise decreasing those weights. This changes the probability distribution exploiting the results of the playouts.
    \item The new policy is then used for the following playout, as well as the biases.
    \item This process is repeated for $n_{policies}$, through which the policy is updated at each iteration and used in the next one, whether a better solution is found, in which case, its actions' weights are increased, or the actions' weights of the last best solution are increased. 
\end{enumerate}

\begin{equation}
\label{weights_gnrpa}
    p_{a,s} = \frac{e^{\frac{w_{a}}{\tau} + \beta_{a}}}{\sum_{a'} e^{\frac{w_{a'}}{\tau} + \beta_{a'}}}
\end{equation}

\begin{equation}
\label{weight_update_gnrpa}
    w_{a} \gets w_{a} - \alpha \frac{p_{a,s} - \delta_{aa'}}{\tau}
\end{equation}

If we consider a GNRPA with bias-free policy, we get a simple Nested Rollout Policy Adaptation (NRPA) algorithm \cite{rosin2011nested,cazenave2021generalized} which follows the same procedure. For a level 2 GNRPA, we consider \emph{two levels of nesting}. At the first level, a policy is initialized with the biases. Then, at the second level, a GNRPA is conducted as described above for $n_{nested\_policies}$ iterations using a copy of the initial policy. At the end of these iterations, the best sequence is used to update the weights of the first level policy, regardless of the weights of the nested policy resulting at the second level. We then use this recently updated policy of level 1 for a second run of a lower level GNRPA. From this run, we also take the best sequence and update the higher level policy, and so on. Using more levels of nesting increases the computation time exponentially, with GNRPA depending on the depth of the search tree and the number of nested policies.

\begin{figure}
\centering
\begin{tikzpicture}
\begin{scope}[every node/.style={circle,thick,draw}]
    \node (A) at (3,0) {Root};
    \node (B) at (1,-2) {.};
    \node (C) at (3,-2) {.};
    \node (D) at (5,-2) {.};
    \node (E) at (0.5,-4) {.};
    \node (F) at (1.5,-4) {.};
    \node (G) at (2.5,-4) {.};
    \node (H) at (3.5,-4) {.};
    \node (I) at (5,-4) {.};

    \node (J) at (0.75,-6) {.};
    \node (K) at (2.25,-6) {.};

    \node (Z) at (0.75,-8) {Leaf};
    \node[rectangle] (S) at (4,-8) {Score};

    \node (A2) at (10,0) {Root};
    \node (B2) at (8,-2) {.};
    \node (C2) at (10,-2) {.};
    \node (D2) at (12,-2) {.};
    \node (E2) at (7.25,-4) {.};
    \node (F2) at (8.75,-4) {.};
    \node (G2) at (9.5,-4) {.};
    \node (H2) at (10.5,-4) {.};
    \node (I2) at (12,-4) {.};
    \node (J2) at (7.75,-6) {.};
    \node (K2) at (9.75,-6) {.};

    \node (Z2) at (7.75,-8) {Leaf};
    \node[rectangle] (S2) at (11,-8) {Score};

\end{scope}

\begin{scope}[>={Stealth[black]},
              every node/.style={fill=white,circle},
              every edge/.style={draw=black,very thick}]
    \path [->] (A) edge[draw=green] node {$5$} (B);
    \path [->] (A) edge node {$3$} (C);
    \path [->] (A) edge node {$1$} (D);
    
    \path [->] (B) edge node {$5$} (E);
    \path [->] (B) edge[draw=green] node {$5$} (F);

    \path [->] (C) edge node {$5$} (G);
    \path [->] (C) edge node {$6$} (H);
    
    \path [->] (D) edge node {$4$} (I);

    \path [->] (F) edge[draw=green] node {$4$} (J);
    \path [->] (F) edge node {$2$} (K);

    \path [->] (J) edge[draw=green] node {...} (Z);
    \path [->] (Z) edge[draw=orange] (S);

    \path [->] (A2) edge[draw=red] node {$\textbf{5.87}$} (B2);
    \path [->] (A2) edge[draw=blue] node {$\textbf{2.13}$} (C2);
    
    \path [->] (A2) edge[draw=blue] node {$\textbf{0.13}$} (D2);
    
    \path [->] (B2) edge[draw=blue] node {$\textbf{4.5}$} (E2);
    \path [->] (B2) edge[draw=red] node {$\textbf{5.5}$} (F2);
    \path [->] (C2) edge[draw=black] node {$5$} (G2);
    \path [->] (C2) edge[draw=black] node {$6$} (H2);
    \path [->] (D2) edge[draw=black] node {$4$} (I2);
    \path[->] (F2) edge[draw=red] node {$\textbf{4.88}$} (J2);
    \path[->] (F2) edge[draw=blue] node {$\textbf{1.12}$} (K2);
    
    \path [->] (J2) edge[draw=red] node {...} (Z2);
    \path [->] (Z2) edge[draw=orange] (S2);
    
\end{scope}
\end{tikzpicture}
\caption{Representation of a Nested Rollout Policy Adaptation, also compatible with GNRPA. The figure on the left represents a rollout, and arcs in green represent the chosen actions at each step. In this figure, the sequence obtained is the best one found so far. We use it to adapt the policy, as described in \hyperref[gnrpa]{section \ref*{gnrpa}}. Otherwise, we use the last best sequence for the policy update. Then we run the next rollout shown on the right. The values in bold are the ones that were updated, the incremented ones have their arcs in red, belonging to the best sequence and the decremented ones are in blue.}
\end{figure}
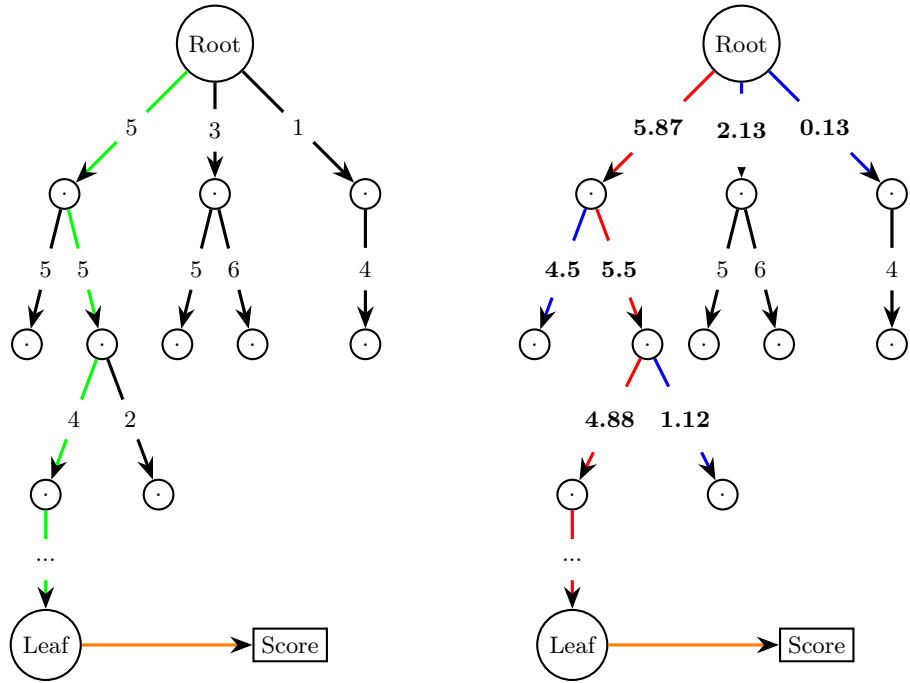

%

\subsection{Motivation}

Improving the total makespan of an industrial assembly line has a clear impact on the productivity and cost-efficiency, which highlights the importance of FJSSP. Exact methods are effective in providing optimal or near-optimal results \cite{meng2020mixed,zhou1996constraint,da2022industrial}. However, they can struggle in some extensions of the FJSSP where operations can be added dynamically \cite{rajabinasab2011dynamic,chang2022deep}, transportation times between machines are considered \cite{karimi2017scheduling} or when uncertainties arise regarding processing times \cite{jamrus2017hybrid}, machine capacity \cite{amiri2019multi}, machine failures or cost fluctuation \cite{meilanitasari2021review}. 

MCTS-based methods have proven their efficacy in dealing with several problems and provide state-of-the-art results on games like Go \cite{silver2016mastering} and Chess \cite{clark2021deep} or biology applications like inverse RNA folding \cite{yang2017rna} as well as energy systems, graph navigation, and, most importantly for us, combinatorial optimization problems \cite{kemmerling2024beyond}. These methods require little computational and memory resources, rendering them compatible with embedded applications.

To the best of our knowledge, only a few attempts use MCTS to solve the FJSSP or its variants \cite{saqlain2023monte,wu2013multi,li2021effective,lu2016dynamic} and they show promise. Yet, they only rely on classic MCTS and do not leverage the more effective potential of its more recent variants such as GNRPA shown in other works \cite{rosin2011nested,cazenave2009nested,sentuc2021generalized} where the information is mostly held by the choice of action independently from the state. We do highlight that this characteristic does not generalize to all problems such as navigation and some games where states contain most of the information with a limited set of possible actions in every state. In addition, previous research only focused on generated or private instances related to industrial needs or on a small number of public benchmarks. Thus, we investigate the performance of these methods in this particular setting on more public benchmark datasets.

We have initially conducted a \hyperref[relevance]{relevance study}, available in the \hyperref[appendix]{appendix}, to assess the usefulness of using such algorithms on this particular problem. Furthermore, we propose a novel variant that relies on the heuristic used to compute the biases. Our intuition is that adapting the biases throughout the playouts will provide a more adequate prior for the considered instance. Although it can still converge to a local minimum, this minimum might prove to be better than what previous variations of the MCTS can provide.

\subsection{Disclaimer}

Our objective is not necessarily to outperform state-of-the-art methods, given that CP-models are more effective than search-based ones here. Indeed, the results available in the literature and to which we are comparing ours are given mostly by CP-based models \cite{dauzere2024flexible}. However, we aim to improve the performance of existing MCTS algorithms and showcase it on an academic problem with several benchmarks.

\section{The Adaptive Bias Generalized Nested Rollout Policy Adaptation (ABGNRPA)}
\label{abgnrpa}

Our main contribution is an enhancement of GNRPA, called ABGNRPA. Compared to GNRPA, after initializing the biases in the same way, we \textbf{update them at each step of the playout}. This forces the bias to be dynamic and to rely on the results of the exploration for a better estimation of the potential of the trajectory. This differs from the update of the weights of the policy which happens at the end of the trajectory using a reward, \emph{i.e} the makespan, whereas biases are updated at every step of every playout.

\subsection*{Application to FJSSP}

The GNRPA and ABGNRPA biases are initialized using the same heuristic called \emph{Earliest Ending Time}. For every action, we compute the minimal starting time possible of processing the operation, given the current state, and add the duration corresponding to the couple (operation, machine). The operations finishing earlier are given a higher value as in \hyperref[bias_initialization]{equation \ref*{bias_initialization}}. We also normalize the biases taking into account all possible actions, see \hyperref[bias_normalization]{equation \ref*{bias_normalization}}. The choice of this heuristic is based on experimental results as it gives a better overall performance compared to other ones, as per the appendix, \hyperref[compare_heuristics]{table \ref*{compare_heuristics}}. The superiority of this heuristic appears to come from the facts that it is optimistic and captures some information from the current state. 

The update of the biases, which is specific to ABGNRPA, is based on a FJSSP makespan lower bound ($\mathcal{LB}$). It is shown in \hyperref[weightupdateabgnrpa]{equation \ref{weightupdateabgnrpa}} where $i$ and $j$ refer respectively to the job and operation, $l$ to the machine, $\mathcal{LB}_{ijl}$ is the makespan lower bound value in the current state, $\gamma$  is a learning rate, and $\delta_{ijl}=1$ if assigning operation $j$ of job $i$ on machine $l$ is chosen or $\delta_{ijl} = -\frac{1}{n_{actions}}$ otherwise.

The $\mathcal{LB}$ is the maximum remaining time on all jobs $i \in \{1..n\}$, added to the current makespan at time $t, ~\mathcal{M}_t$. For every job, we compute the sum of the minimal durations of every operation that remains, which are represented by the set $\mathcal{U}=\{ o_{i,j}, i \in \{1..n\} \text{ where } o_{i,j} \text{ has not been assigned to a machine yet}\}$ and take the maximum value obtained, as detailed in \hyperref[minimal_estimate]{equation \ref*{minimal_estimate}}. These updated biases are then used for the following playouts, as shown in \hyperref[biased_policy_playout]{algorithm 1}. We believe we can enlarge the application of this algorithm to other problems given a relevant lower bound function.

\begin{equation}
    \label{bias_initialization}
    EET_{ijl} \gets -(earliest\_start\_time(action_{ijl}, state) + d_{ijl})
\end{equation}

\begin{equation}
    \label{bias_normalization}
    \beta_{ijl} \gets \frac{EET_{ijl}}{\sum_{action} EET_{action}}
\end{equation}

\begin{equation}
    \label{weightupdateabgnrpa}
    \beta_{ijl} \gets \beta_{ijl} + \gamma * \delta_{ijl} * \mathcal{LB}_{ijl}
\end{equation}

\begin{equation}
    \label{minimal_estimate}
    \mathcal{LB}_{ijl} = \mathcal{M}_t + \max_{i' \in \{1..n\}} [ \sum_{o_{i'k} \in ~ \mathcal{U}} \min_{l' \in \{1..m\}}(d_{i'kl'}) ]
\end{equation}

\begin{algorithm}
    \label{biased_policy_playout}
    \caption{Biased Policy Playout for ABGNRPA}
    \begin{algorithmic}
        
        \State \textbf{Input:} state, policy, heuristic\_values, $\gamma$, $\tau$
        \State \textbf{Begin}
        \State list\_of\_moves $\gets $ state.possible\_moves()\; \Comment{\textit{Moves are tuples (job, operation, machine)}}
        \While{list\_of\_moves is not empty}
            \For{move $\in$ list\_of\_moves}
                \State probabilities[move] $ \gets e^{\frac{\text{$\alpha_{move}$}}{\tau} + \text{$\beta_{move}$}}$ \;
                \Comment{\textit{table of $p_{move}$ values} of size $n_{moves}$}
            \EndFor
            \State next\_move $\gets$ \verb|random_weighted_choice|(list\_of\_moves, probabilities)
            \State \text{$\mathcal{LB}$} $\gets$ \verb|LB_function|(next\_move)
            \State state.play(next\_move)
            \For{move $\in$ list\_of\_moves}
                \If{move $=$ next\_move}
                    \State $\delta$ $\gets 1$
                \Else
                    \State $\delta$ $\gets \frac{-1}{\text{size(list\_of\_moves)}}$
                \EndIf
                \State $\beta_{move}\gets\beta_{move}$ + $\gamma~*\delta~*$ \text{$\mathcal{LB}$}
            \EndFor
            \State list\_of\_moves $\gets $ state.possible\_moves()
        \EndWhile
    \State \textbf{End}
    \end{algorithmic}
\end{algorithm}
    
We highlight that our contribution differs from \cite{sentuc2021generalized}. Indeed, for the Capacitated Vehicle Routing Problem with Time Windows (CVRPTW), at each state they take into account the distance to the arrival point, the wait time and the arrival time, making the bias \emph{dynamic}. In our case, we initialize the biases with the heuristic value and we update at each step them using $\mathcal{LB}$ which is computed from the current state instead of recomputing the bias. This means that the bias accumulates modifications every time the action is encountered.

\section{Experiments and results}
\label{results}

\subsection{Model description and implementation}
\label{model}

We construct our model considering the problem as a sequential decision making process by ordering the different operations on the machines using the inherent partial order given by the jobs, and compute the starting time of each of these operations after the allocation. The idea is to visualize the problem as a tree search problem and implement MCTS methods to solve it as explained in the following:

\begin{enumerate}
    \item Since operations in a job have a specific order, we will allocate them following that order and we start by the first operation. The choices for the first action are then the couples containing the first operations of all jobs, \emph{i.e}, the subset $\mathcal{RO} = \{ o_{i,0} ~ , ~ i \in \{1..n\}\}$, and their compatible machines $\mathcal{A} = \{ (o_{i,0}, m_l), , ~ i \in \{1..n\}, l \in \{1..m\}, \text{ 
 if  } o_{i,0} \text{ and } m_l$ are compatible \}.
    \item An operation is chosen, using the policy, and allocated on a compatible machine $m_{l_0}$. The starting time is then $t_{start,i_0,l_0}=0$ and the ending time is $t_{end,i_0,0}=d_{i_0,0,m_{l_0}}$. 
    \item Given that we decided to process the operations following the job order, the only operation from job $i_0$ that can be allocated at this point is $o_{i_0, 1}$. It then replaces $o_{i_0, 0}$ in $\mathcal{RO}$.
    \item We repeat the previous steps, each time removing the chosen operation and replacing it by the following one in the same job until there are no operations left in $\mathcal{RO}$.
    
\end{enumerate}

We implemented the model using \verb|Python|, version 3.9.19. All compared algorithms have been implemented from scratch and use native \verb|Python| functions and class types, which enables a fair comparison of all baseline algorithms described in \hyperref[baseline]{section \ref*{baseline}}.

\subsection{Datasets}
\label{datasets}

Several benchmarking instances are available in the literature \cite{dauzere2024flexible} \footnote{\url{https://github.com/SchedulingLab/fjsp-instances}}. We choose to consider a subset of these instances, with $|\mathcal{M}_{ij}|$ representing the flexibility, \emph{i.e} the number of compatible machines with operation $o_{ij}$:

\begin{itemize}
    \item 10 instances from Brandimarte's \cite{brandimarte1993routing}: $n=10..20$, $m=4..15$, $|\mathcal{J}_i|=3..15$, $d_{ijl}=1..20$, $|\mathcal{M}_{ij}|=2..6$.
    \item 4 instances from Kacem's \cite{kacem2002pareto}: $n=4..10$, $m=5..10$, $\sum_i |\mathcal{J}_i| =12..56$, $|\mathcal{M}_{ij}|=m$, $d_{ijl}=1..10$ with some outliers such as 54 in \emph{k1}
    \item 66x3 instances from Hurink's \cite{hurink1994tabu} (\emph{edata, rdata, vdata}) derived from \emph{sdata-instances}. With $n=6..30$, $m=5..15$, flexibility varies as:
        \begin{itemize}
            \item \emph{edata}: average is $1.15$, max is $3$ (few operations have flexibility)
            \item \emph{rdata}: average is $2$, max is $3$ (most operations have flexibility)
            \item \emph{vdata}: average is $\frac{1}{2}m$, max is $\frac{4}{5}m$ (all operations have flexibility).
        \end{itemize}
    
\end{itemize}

All these descriptions are taken from \cite{behnke2012test}. These instances vary in size and flexibility, giving a variety of instance types to benchmark our approach. In addition, optimal values were found for all instances, except \emph{mk10} from \cite{brandimarte1993routing}.

\subsection{Baseline algorithms}
\label{baseline}

We compare our algorithm to some variations of MCTS algorithms: classic MCTS, Nested Monte Carlo Tree Search (NMCTS) \cite{cazenave2009nested}, Nested Rollout Policy Adaptation (NRPA) \cite{rosin2011nested}, Generalized Nested Rollout Policy Adaptation (GNRPA) \cite{cazenave2021generalized} as well as other devised ones:

\begin{itemize}
    \item \emph{Heuristic-Based NMCTS} (\textbf{HBNMCTS}): the playouts are biased instead of uniformly random.
    \item \emph{Bias Initialized NRPA} (\textbf{BINRPA}): the weights of actions are initialized using the biases provided by the heuristic instead of a uniform distribution.
    \item \emph{Randomized Greedy} (\textbf{RandGreedy}): no exploration is done, only a weighted random playout following the values of a heuristic.
\end{itemize}

%
All these algorithms are run for level 1, as explained in \hyperref[gnrpa]{section \ref*{gnrpa}} for GNRPA-based approaches and as described in \cite{cazenave2009nested} for NMCTS-based approaches. \emph{Randomized Greedy} is the only exception as it only plays one rollout and does not conduct an exploration in the tree. We allow a budget of $60$ seconds for every algorithm, running on $100$ independent iterations.

\subsection{Experimental Results}

\vspace*{-5mm}

\begin{table}[]
    \centering
    \begin{tabular}{|c|c|c|}
        \hline
        \multirow{2}{*}{Algorithm} & Best value & UB reached \\
         & frequency & count \\
        \hline
        NRPA & 1 & 1 \\
        \hline
        MCTS & 2 & 2 \\
        \hline 
        NMCTS & 0 & 0 \\
        \hline 
        HBNMCTS & 32 & 3 \\
        \hline
        BINRPA & 1 & 1 \\
        \hline
        RandGreedy & 1 & 1 \\
        \hline
        GNRPA & 72 & 4 \\
        \hline
        ABGNRPA & \textbf{126} & \textbf{8} \\
        \hline
    \end{tabular}
    \begin{tabular}{|c|c|c|}
        \hline
        \multirow{2}{*}{Average STD} & \multicolumn{2}{c|}{Gap to UB} \\
        \cline{2-3}
         & Min makespan & Mean makespan \\
        \hline
        7.59\% & 54.21\% & 83.16\%\\ 
        \hline
        2.88\% & 52.93\% & 66.66\% \\
        \hline
        5.01\% & 61.00\% & 83.00\% \\
        \hline
        3.05\% & 15.52\% & 23.91\% \\
        \hline
        2.94\% & 52.71\% & 66.88\% \\
        \hline
        9.61\% & 53.33\% & 110.45\% \\
        \hline
        2.71\% & 14.27\% & 18.79\% \\
        \hline
        \textbf{1.43\%} & \textbf{12.08\%} & \textbf{17.84\%} \\
        \hline
        
    \end{tabular}
    \caption{The table on the left showcases the number of times each algorithm obtains the best makespan compared to the others, and the number of times it finds the known Upper Bounds (UB) from literature \cite{dauzere2024flexible,graviers2025updating}. The higher the value the better. The table on the right shows the average STandard Deviation (STD) of the makespan for each algorithm and the average gap on all instances between the upper bound and minimal and mean makespan values found. In this table, the lower the better.}
    \label{tab:gap_to_UB}
\end{table}

\vspace*{-15mm}

\begin{table}[]
    \centering
    \begin{tabular}{|c|c|c|c|c|c|}
        \hline
        Instanes groups & \emph{edata} & \emph{rdata} & \emph{vdata} & Kacem & Brandimarte \\
        \hline
        Max deviation & 31.51\% & 32.14 \% & 34.14\% &  28.06\% & 30.07\% \\
        \hline
        Min deviation & 7.59\% & 7.20\% & 5.30\% & 13.38\% & 10.19\% \\
        \hline
        Average deviation & 15.27\% & 14.59\% & 13.28\% & 20.86\% & 17.31\% \\
        \hline
    \end{tabular}
    \caption{Bias deviation between the start and the end of an ABGNRPA run in percentages, aggregated on instances' datasets from \cite{hurink1994tabu,kacem2002pareto,brandimarte1993routing}}
    \label{tab:bias_differnece}
\end{table}

An aggregation of the results obtained are available in table \ref{tab:gap_to_UB}. Exhaustive results are shown in the \hyperref[appendix]{appendix} as per-instance results (minimal and average makespans and standard deviation) with averages ranking, all in \hyperref[exhaustive_results]{section \ref*{exhaustive_results}}. We also provide further details on nesting levels' impact in \hyperref[levels]{section \ref*{levels}}.

\subsection{Observations and discussion}
\label{discussion}

In table \ref{tab:gap_to_UB}, we observe that, on \textbf{126 out of 212} instances, ABGNRPA performs better than the other algorithms. It gets closest to the optimal and even reaches it on \textbf{8 instances}, and GNRPA and HBNMCTS are right behind it. This shows that the addition of biases helps to improve the initial performance of NRPA. However, GNRPA does not appear to sufficiently exploit the information that can be extracted from the problem structure. The better performance of ABGNRPA implies that the adaptive bias feature abstracts additional information helping it get a better bias estimation. This is also highlighted by its average performance and stability. We can conjecture that the biases are \emph{evolving} to become more compatible with the problem at hand and that this evolution, which occurs during the playouts, helps capture information before its end and capitalize on it to improve the search.

We compute the bias deviation between the start and the end of a ABGNRPA run. Aggregated results are in table \ref{tab:bias_differnece}, and per-instance results are available in \hyperref[bias_deviation_analysis]{section \ref*{bias_deviation_analysis}} of the appendix. Varying between instances, this behavior can explain the better performance of ABGNRPA and indicates that the adaptation of the biases leads to a better representation of the instance's characteristics. However, contrary to other existing applications of GNRPA \cite{dang2023warm,cazenave24ppsn}, increasing the nesting level does not improve performance for GNRPA variants. Our hypothesis is that these algorithms, with this specific model, stagnate around a local minimum.

The last point suggests that further training is rather useless as it would make the policy more deterministic. This hinders the possibility of further exploration beyond the visited branches. To counter this, we can consider using Limited Repetitions \cite{cazenave2024limited}, a combination of different heuristics with a learning scheme for the weights on these heuristics from which the bias is constructed \cite{sentuc2023learning}, or a nested bias system. Another possible improvement is the consideration of a state-action-based policy, and not only an action-based policy, which may resemble a Q-learning scheme \cite{watkins1992q}. For the $\mathcal{LB}$ function, we have tested the formulation mentioned in \hyperref[minimal_estimate]{equation \ref*{minimal_estimate}} but we can also try others that may be more expressive.

\section{Conclusion}
\label{conclusion}

Through this work, we compared the performance of our algorithm, the ABGNRPA, to that of other MCTS variations on the FJSSP and concluded that we have indeed improved the results. However, many parameters are to be considered, such as the heuristic function, the $\mathcal{LB}$ estimation, and the hyperparameters of the algorithm. The bias deviation, presented in table \ref{tab:bias_differnece}, seems to characterize the GNRPA family adaptability. This suggests the need for further investigation to fully explore its potential to handle larger and different problems. Improving the scope and extent of ABGNRPA may prove useful when tackling other planning and scheduling applications. Taking into account different factors that hinder the prowess of CP-based approaches, we can foresee the relevance of our approach in bypassing these obstacles with its stochastic nature.

%
%
%
%

\clearpage

\small{
\bibliographystyle{splncs04}
\bibliography{bibliography}
}

\clearpage

\appendix
\label{appendix}

\chapter*{Appendix}

\section{Relevance study}
\label{relevance}

\subsection{Description}

This section aims at studying the relevance of using MCTS and its variants on the Flexible Job Shop Scheduling Problem. The intuition is, as proposed in \cite{eliahou2013investigating}, that the average makespan of random playouts in a branch is expected to provide a hint on how promising this branch is to explore. Other metrics could be explored but are not considered in this study. In theory, if we consider $X$ the random variable representing the makespan of a playout, a good estimator of the expected value $\mathbf{E}(X)$ is the mean of the playouts $\bar{X}$. We devide the search tree into distinct branches, where in our case, a branch of the search tree starts by assigning sequentially two operations on machines and then all the possible outcomes after these first two steps. Then, we fully explore each branch and compute the average and minimal makespans for this given branch. We then evaluate the correlation between the mean and minimum given by all branches to assess whether the mean makespan is a good indicator of the branch’s ability to provide a good solution. Studying the standard deviation did not provide substantial information except the fact that the deviation of the larger makespans compared to the mean is greater than that of the smaller ones.

\subsection{Test instances}

We consider a small subset of the instances described in \hyperref[datasets]{section \ref*{datasets}} as well as others available in the literature \cite{behnke2012test}. For each group of datasets, we choose a random instance and reduce its content to explore the search tree fully and reasonably quickly. \hyperref[tab:relevance_instances]{Table \ref*{tab:relevance_instances}} summarizes the size of each instance considered.

\begin{table}[]
    \centering
    \begin{tabular}{| c | c | c | c | c | c |}
        \hline
        \textbf{Reduced instance} & $n$ & $m$ & $n_i$ & Flexibility & $d_{j,i}$ \\
        \hline
        \textbf{mt10xxx} & 4 & 13 & 2..3 & 1..3 &  10..98 \\
        \textbf{lar03\_1} & 4 & 60 & 2..3 & 1..2 & 10..30 \\
        \textbf{mk03} & 4 & 8 & 2..3 & 2..4 & 1..19 \\
        \textbf{05a} & 3 & 5 & 4..5 & 1..3 & 18..84 \\
        \textbf{mfjs04} & 4 & 7 & 2..3 & 2..3 & 65..247 \\
        \textbf{abz9} & 3 & 15 & 3..5 & 2..3 & 11..40 \\
        \textbf{k3} & 4 & 10 & 1..3 & 1..7 & 1..10 \\
        \hline
    \end{tabular}
    \caption{Description of the instances considered for the relevance study}
    \label{tab:relevance_instances}
\end{table}

\subsection{Experimental results}

\begin{figure}
    \centering
    \includegraphics[width=13cm]{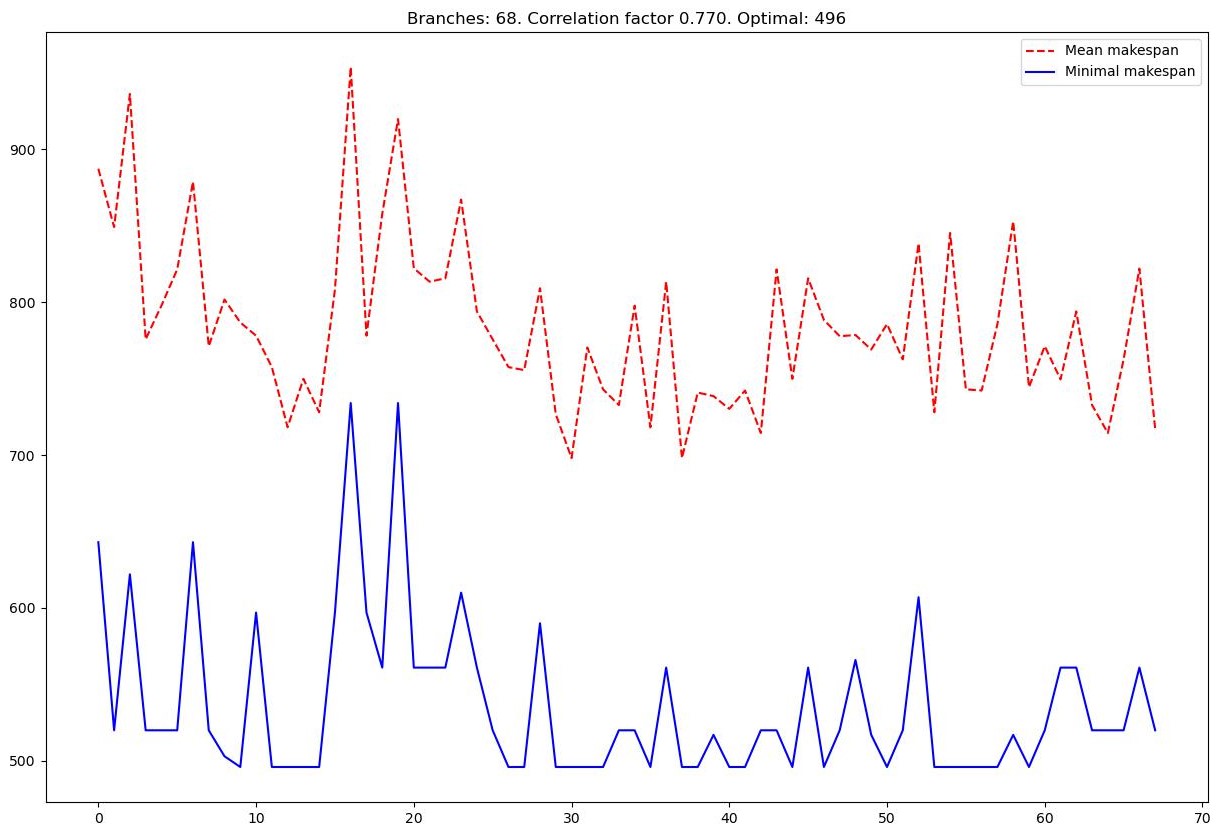}
    \caption{Correlation between mean and minimal makespans for each branch for the reduced mfjs04 instance.}
    \label{fig:correlation_fattahi}
\end{figure}

The optimal values and correlation between minimal and average makespans computed for each branch are shown in \hyperref[tab:relevance_results]{table \ref*{tab:relevance_results}}. An example of how the mean and minimal makespan vary in each branch is shown in \hyperref[fig:correlation_fattahi]{figure \ref*{fig:correlation_fattahi}}.

\begin{table}[]
    \centering
    \begin{tabular}{|c|c|c|c|c|c|c|c|}
        \hline
        \textbf{Reduced Instance} & \textbf{mt10xxx} & \textbf{lar03\_1} & \textbf{mk03} & \textbf{05a} & \textbf{mfjs04} & \textbf{abz9} & \textbf{k3} \\
        \hline
        \textbf{Correlation} & 0.643 & 0.685 & 0.709 & 0.438 & 0.770 & 0.749 & 0.887  \\
        \textbf{Optimal} & 197 & 67 & 21 & 305 & 496 & 170 & 7  \\
        \textbf{Number of branches} & 32 & 60 & 111 & 42 & 68 & 56 & 194  \\
        \hline
    \end{tabular}
    \caption{Correlation values, optimal values and number of branches for the instances described in \hyperref[tab:relevance_instances]{table \ref*{tab:relevance_instances}}}.
    \label{tab:relevance_results}
\end{table}

We notice that on most instances, the correlation factor is relatively high, which is a good indicator of the validity of using MCTS methods on this type of problems. The low correlation value for the simplified instance of \emph{05a} is a result of the fact that we can find a near-optimal solution in a high number of branches, while the mean value of the makespan differs between branches.

\section{Parameters of the experimental procedure}
\label{experimental_procedure}

\subsection{The choice of the heuristic for bias computation}

When using GNRPA as a baseline to compare our algorithm's performance, we faced the task of choosing the appropriate heuristic value for computing the biases. We came up thus with a list of choices, some of them were described in the literature as usable for JSSP instances and others were of our own creation:

\begin{itemize}
    \item \emph{Shortest Processing Time} (\textbf{SPT}): \emph{i.e}, the shorter the processing time of the combination of (operation, machine), the higher the bias value
    \item \emph{Longest Processing Time} (\textbf{LPT)}): we prioritize the couples (operation, machine) with the longest processing time.
    \item \emph{Least Remaining Processing Time} (\textbf{LRPT}): we prioritize the operation of the job that has the least remaining time (we consider the lowest processing time for each of the remaining operations for this job).
    \item \emph{Most Remaining Processing Time} (\textbf{MRPT}): the prioritization process is the opposite of the last heuristic.
    \item \emph{Earliest Ending Time} (\textbf{EET}): we prioritize the operations that would have the earliest ending time. For that, we compute the feasible start time with the addition of the processing time of the machine of the action, given the current state.
    \item \emph{Latest Ending Time} (\textbf{LET}): operations with the latest ending time are prioritized.
\end{itemize}

We compare the performance of the GNRPA for each of these heuristics on some of the instances in the body of the article under the same conditions. The results shown in \hyperref[compare_heuristics]{table \ref*{compare_heuristics}} show that the SPT and EET heuristics perform better than the pessimistic ones. In particular, EET provides the best results. This shows how important the choice of the heuristic is. When we use the same comparison for ABGNRPA, we observe similar behavior. These empirical results are the basis of the use of this heuristic in our experimental procedure.

\begin{table}[]
    \centering
    \begin{tabular}{|c|c|c|c|c|c|c|}
        \hline
        Instance & LRPT & MRPT & SPT & LPT & EET & LET \\
        \hline
        k1 & 13 & 14 & 13 & 15 & \textbf{12} & 14 \\
        k2 & 29 & 29 & 29 & 30 & \textbf{10} & 28 \\
        k3 & 21 & 23 & 24 & 26 & \textbf{8} & 20 \\
        k4 & 46 & 47 & 46 & 43 & \textbf{12} & 47 \\
        mk01 & 61 & 61 & 62 & 61 & \textbf{44} & 61 \\
        mk02 & 59 & 59 & 59 & 58 & \textbf{35} & 59 \\
        mk03 & 336 & 333 & 343 & 349 & \textbf{222} & 338 \\
        mk04 & 105 & 100 & 97 & 109 & \textbf{71} & 101 \\
        mk05. & 239 & 234 & 237 & 238 & \textbf{184} & 238 \\
        mk06 & 174 & 176 & 167 & 180 & \textbf{70} & 172 \\
        mk07 & 262 & 258 & 255 & 246 & \textbf{169} & 250 \\
        mk08 & 664 & 653 & 655 & 657 & \textbf{525} & 658 \\
        mk09 & 588 & 571 & 600 & 592 & \textbf{384} & 590 \\
        mk10 & 490 & 513 & 516 & 524 & \textbf{305} & 515 \\
        \hline    
        
    \end{tabular}
    \caption{Comparison of the performance of GNRPA (minimal makespan on 100 runs of the algorithm) using different heuristics}
    \label{compare_heuristics}
\end{table}

\subsection{ABGNRPA performance with different heuristics}

It seems essential to validate the compatibility between our algorithm and the heuristics we chose to test. While the EET heuristic provides the best results for the GNRPA as well as the ABGNRPA, as mentioned above, the better performance of the ABGNRPA compared to the other algorithms does not necessarily hold true when using a different heuristic, as shown in table \ref{gnrpa_abgnrpa_heuristics}.

\begin{table}
    \raggedleft
    \begin{tabular}{llll}
        \begin{tabular}{|c|c|c|}
            \hline
            \multirow{2}{*}{Instance} & \multicolumn{2}{|c|}{MRPT}\\
            \cline{2-3}
            & GNRPA & ABGNRPA \\
            \hline
            k1 & 14 & 67 \\
            k2 & 29 & 191 \\
            k3 & 23 & 83 \\
            k4 & 47 & 180 \\
            mk01 & 61 & 76 \\
            mk02 & 59 & 75 \\
            mk03 & 333 & 491 \\
            mk04 & 100 & 140 \\
            mk05 & 234 & 218 \\
            mk06 & 176 & 220 \\
            mk07 & 258 & 315 \\
            mk08 & 653 & 679 \\
            mk09 & 571 & 680 \\
            mk10 & 513 & 599 \\
            \hline
        \end{tabular}
    
        \begin{tabular}{|c|c|}
            \hline
            \multicolumn{2}{|c|}{LRPT}\\
            \hline
            GNRPA & ABGNRPA \\
            \hline
            13 & 12 \\
            29 & 15 \\
            21 & 9 \\
            46 & 26 \\
            61 & 84 \\
            59 & 80 \\
            336 & 585 \\
            105 & 216 \\
            239 & 492 \\
            174 & 272 \\
            262 & 360 \\
            664 & 1866 \\
            588 & 1554 \\
            490 & 1262 \\
            \hline
     
        \end{tabular}
        \begin{tabular}{|c|c|}
            \hline
            \multicolumn{2}{|c|}{LPT}\\
            \hline
            GNRPA & ABGNRPA \\
            \hline
            15 & 66 \\
            30 & 172 \\
            26 & 81 \\
            43 & 223 \\
            61 & 75 \\
            58 & 84 \\
            349 & 412 \\
            109 & 132 \\
            238 & 226 \\
            180 & 235 \\
            246 & 313 \\
            657 & 645 \\
            592 & 518 \\
            524 & 455 \\
            \hline
     
        \end{tabular}

        \begin{tabular}{|c|c|}
            \hline
            \multicolumn{2}{|c|}{SPT}\\
            \hline
            GNRPA & ABGNRPA \\
            \hline
            13 & 11 \\
            29 & 14 \\
            24 & 8 \\
            46 & 16 \\
            62 & 51 \\
            59 & 38 \\
            343 & 233 \\
            97 & 101 \\
            237 & 210 \\
            167 & 100 \\
            255 & 171 \\
            655 & 571 \\
            600 & 418 \\
            516 & 343 \\
            \hline
            
        \end{tabular}
    \end{tabular}
    \caption{Comparison of GNRPA and ABGNRPA using different heuristics}
    \label{gnrpa_abgnrpa_heuristics}
\end{table}

We observe that the improvement in performance holds for the SPT heuristic in the same way as for EET, while being unfavorable for the other ones, especially when it comes to pessimistic heuristics.

\subsection{Correlation between the bias values and the policy values}
\label{correlation}

For both GNRPA and ABGNRPA, we observe better performance in general compared to the \emph{Randomized Greedy} baseline. This justifies the thought that adding a policy to the biases improves performance. We then decide to verify the correlation between the policy weights and the biases at the end of the running of each algorithm, and we obtained similar results for ABGNRPA and GNRPA. The \textbf{average correlation is 0.2}, which means that the heuristic serves as a good start for these algorithms but they learn to deviate from the initial weights. In the same sense, this also justifies the better performance of ABGNRPA relatively to that of GNRPA as it shifts the biases as well.

\subsection{The effect of increasing nesting levels}
\label{levels}

\begin{table}[]
    \centering
    \begin{tabular}{|c|cc|cc|}
        \hline
        & \multicolumn{2}{|c|}{ABGNRPA} & \multicolumn{2}{|c|}{GNRPA} \\

        Instance & Level 1 & Level 2 & Level 1 & Level 2 \\
        \hline
        k1 & 11 & 11 & 12 & 12 \\
        k2 & 11 & 11 & 13 & 13 \\
        k3 & 7 & 7 & 8 & 8 \\
        k4 & 12 & 12 & 12 & 12 \\
        mk01 & 44 & 43 & 45 & 44 \\
        mk02 & 32 & 32 & 34 & 35 \\
        mk03 & 226 & 232 & 220 & 222 \\
        mk04 & 67 & 67 & 70 & 71 \\
        mk05 & 181 & 181 & 183 & 184 \\
        mk06 & 66 & 65 & 69 & 70 \\
        mk07 & 164 & 164 & 165 & 169 \\
        mk08 & 527 & 529 & 528 & 525 \\
        mk09 & 347 & 351 & 386 & 384 \\
        mk10 & 279 & 278 & 295 & 305 \\
        \hline
    \end{tabular}
    \caption{Comparison of minimal makespan obtained by GNRPA-based algorithms with one or two levels of nesting}
    \label{tab:levels}
\end{table}

We observe in table \ref{tab:levels} that increasing the level does not improve the performance of GNRPA and ABGNRPA. This is not the case for NRPA and NMCTS as shown in table \ref{tab:levels_nrpa_nmcts}.

\begin{table}[]
    \centering
    \begin{tabular}{|c|cc|cc|}
        \hline
       & \multicolumn{2}{|c|}{NMCTS} & \multicolumn{2}{|c|}{NRPA} \\
        Instance &  Level 1 & Level 2 & Level 1 & Level 2 \\
        \hline
        k1 & 11 & 11 & 12 & 11 \\
        k2 & 25 & 18 & 25 & 16 \\
        k3 & 18 & 14 & 21 & 13 \\
        k4 & 32 & 29 & 38 & 28 \\
        \hline

    \end{tabular}
    \caption{Comparison of minimal makespan obtained with NMCTS and NRPA with nesting of levels 1 and 2}
    \label{tab:levels_nrpa_nmcts}
\end{table}

\subsection{The use of 2-opt permutations}
\label{two-opt}

Since we are constructing a sequence through the resolution of our problem, we figured it may be interesting to try \verb|2-opt| permutations on this sequence at the end of a playout to see if we can improve the makespan through local search. In short, these permutations do improve the performance of all algorithms except ABGNRPA, but do not exceed its performance, which strengthens the idea that we are saturating the performance of the chosen tree search methods on this problem.

\section{Detailed performance metrics}
\label{exhaustive_results}

\subsection{Per-instance minimal makespan}

\begin{table}[!h]
    \tiny
    \centering
\begin{tabular}{|c|c|c|c|c|c|c|c|c|c|}
        \hline
        Instance & NRPA & MCTS & NMCTS & HBNMCTS & BINRPA & RandGreedy & GNRPA & ABGNRPA & Upper bound \\
        \hline
         abz5 & 1534 & 1503 & 1671 & 1218 & 1471 & 1506 & \textbf{1212} & \textbf{1212} & 1167* \\
         abz6 & 1149 & 1172 & 1274 & 1011 & 1198 & 1237 & 997 & \textbf{978} & 925* \\
         abz7 & 1191 & 1078 & 1128 & 754 & 1083 & 1075 & 778 & \textbf{742} & 610 \\
         abz8 & 1262 & 1138 & 1210 & 785 & 1176 & 1127 & 799 & \textbf{776} & 636 \\
         abz9 & 1217 & 1104 & 1161 & 821 & 1123 & 1114 & 837 & \textbf{808} & 644* \\
         car1 & 7127 & 7980 & 8469 & 7422 & 7544 & 7416 & \textbf{6708} & 6726 & 6176* \\
         car2 & 7462 & 7881 & 8161 & 7212 & 7766 & 7534 & \textbf{6534} & 6586 & 6327* \\
\hline
    \end{tabular}
    \caption{Experimental results on Hurink's \cite{hurink1994tabu} \emph{edata}, with a time budget of 60 seconds. * means the value is optimal.}
    \label{tab:edata_results}
\end{table}

\clearpage

\begin{table}[!h]
    \tiny
    \centering
    \begin{tabular}{|c|c|c|c|c|c|c|c|c|c|}
        \hline
        Instance & NRPA & MCTS & NMCTS & HBNMCTS & BINRPA & RandGreedy & GNRPA & ABGNRPA & Upper bound \\
        \hline
         abz5 & 1498 & 1544 & 1713 & 1127 & 1588 & 1557 & \textbf{1039} & 1040 & 954* \\
         abz6 & 1238 & 1168 & 1227 & 906 & 1096 & 1205 & \textbf{857} & 878 & 807* \\
         abz7 & 1237 & 1137 & 1137 & 684 & 1134 & 1134 & 759 & \textbf{651} & 522 \\
         abz8 & 1282 & 1171 & 1218 & 693 & 1156 & 1121 & 761 & \textbf{684} & 535 \\
         abz9 & 1254 & 1166 & 1181 & 717 & 1137 & 1137 & 780 & \textbf{714} & 536 \\
         car1 & 6218 & 6590 & 6949 & 5940 & 6494 & 6077 & \textbf{5400} & 5455 & 5034* \\
         car2 & 6751 & 7240 & 7257 & 6551 & 6979 & 6820 & 6226 & \textbf{6190} & 5985* \\
\hline
    \end{tabular}
    \caption{Experimental results on Hurink's \cite{hurink1994tabu} \emph{rdata}, with a time budget of 60 seconds. * means the value is optimal.}
    \label{tab:rdata_results}
\end{table}

\clearpage

\begin{table}[!h]
    \tiny
    \centering
    \begin{tabular}{|c|c|c|c|c|c|c|c|c|c|}
         \hline
         Instance & NRPA & MCTS & NMCTS & HBNMCTS & BINRPA & RandGreedy & GNRPA & ABGNRPA & Upper bound \\
         \hline
         abz5 & 1473 & 1485 & 1610 & 899 & 1495 & 1493 & 972 & \textbf{893} & 859* \\
         abz6 & 1195 & 1177 & 1246 & 786 & 1207 & 1226 & 808 & \textbf{772} & 742* \\
         abz7 & 1188 & 1117 & 1114 & 612 & 1111 & 1090 & 645 & \textbf{598} & 492* \\
         abz8 & 1218 & 1136 & 1083 & \textbf{610} & 1145 & 1167 & 712 & \textbf{610} & 507 \\
         abz9 & 1216 & 1122 & 1106 & \textbf{621} & 1088 & 1126 & 700 & 622 & 497* \\
         car1 & 6248 & 6294 & 6830 & 5528 & 6135 & 5689 & 5460 & \textbf{5384} & 5005* \\
         car2 & 6564 & 6657 & 7016 & 6201 & 6829 & 6360 & \textbf{6084} & 6284 & 5929* \\
\hline
    \end{tabular}
    \caption{Experimental results on Hurink's \cite{hurink1994tabu} \emph{vdata}, with a time budget of 60 seconds. * means the value is optimal \cite{dauzere2024flexible,graviers2025updatinglowerupperbounds}.}
    \label{tab:vdata_results}
\end{table}

\clearpage

\begin{table}[!h]
\begin{adjustbox}{width=\linewidth,center}
    \begin{tabular}{|c|c|c|c|c|c|c|c|c|c|}
        \hline
        Instance & NRPA & MCTS & NMCTS & HBNMCTS & BINRPA & RandGreedy & GNRPA & ABGNRPA & Upper bound\\
        \hline
        k1 & 12 & \textbf{11*} & 12 & \textbf{11*} & 12 & 12 & \textbf{11*} & \textbf{11*} & 11* \\
        k2 & 24 & 25 & 27 & 13 & 25 & 28 & \textbf{11*} & \textbf{11*} & 11* \\
        k3 & 18 & 22 & 24 & 7 & 20 & 22 & \textbf{7*} & \textbf{7*} & 7* \\
        k4 & 37 & 39 & 42 & 13 & 41 & 46 & \textbf{12} & \textbf{12} & 11* \\
        mk01 & 55 & 58 & 61 & 44 & 57 & 62 & \textbf{43} & 44 & 40* \\
        mk02 & 48 & 51 & 59 & 34 & 50 & 55 & 32 & \textbf{31} & 26* \\
        mk03 & 330 & 338 & 352 & 238 & 331 & 339 & 222 & \textbf{213} & 204* \\
        mk04 & 95 & 94 & 106 & 72 & 93 & 98 & 70 & \textbf{69} & 60* \\
        mk05 & 214 & 225 & 237 & 188 & 222 & 231 & 187 & \textbf{183} & 172* \\
        mk06 & 170 & 163 & 164 & 71 & 164 & 177 & \textbf{68} & \textbf{68} & 57* \\
        mk07 & 236 & 251 & 255 & 170 & 241 & 254 & 163 & \textbf{158} & 139* \\
        mk08 & 701 & 651 & 675 & 547 & 663 & 662 & 526 & \textbf{523*} & 523* \\
        mk09 & 614 & 595 & 580 & 358 & 559 & 582 & 390 & \textbf{356} & 307* \\
        mk10 & 496 & 509 & 503 & 303 & 506 & 504 & 310 & \textbf{267} & 193 \\
         \hline

    \end{tabular}
    \end{adjustbox}
    \caption{Experimental results on Kacem's \cite{kacem2002pareto} and Brandimarte's \cite{brandimarte1993routing} instances minimum makespan on 100 iterations of each algorithm with a time budget of 60 seconds for each iteration. * means the value is optimal.}
    \label{tab:kacem_brandimarte_results}
\end{table}

\subsection{Per-instance average makespan}

\begin{table}[]
    \centering
    \begin{tabular}{|c|c|c|c|c|c|c|c|}
        \hline
        NRPA & MCTS & NMCTS & HBNMCTS & BINRPA & RandGreedy & GNRPA & ABGNRPA \\
        \hline
        6.43 & 4.40 & 6.70 & 2.78 & 4.55 & 7.86 & 1.68 & \textbf{1.61} \\
        \hline
    \end{tabular}
    \caption{Average ranking of tested algorithms on the mean makespan obtained on 100 iterations with a budget of 60 seconds on Hurink's, Kacem's, and Brandimarte's instances}
    \label{tab:average_makespan_ranking}
\end{table}

\begin{table}[!h]
    \tiny
    \centering
    \begin{tabular}{|c|c|c|c|c|c|c|c|c|}
        \hline
        Instance & NRPA & MCTS & nmcts & HBNMCTS & BINRPA & RandGreedy & GNRPA & ABGNRPA \\
        \hline
        abz5 & 1835.42 & 1628.19 & 1841.72 & 1309.56 & 1625.47 & 2079.37 & 1247.22 & 1244.25 \\
        abz6 & 1436.91 & 1283.75 & 1489.44 & 1081.84 & 1295.42 & 1657.60 & 1025.44 & 1027.38 \\
        abz7 & 1355.30 & 1156.27 & 1217.89 & 793.33 & 1156.75 & 1365.68 & 803.09 & 774.38 \\
        abz8 & 1433.11 & 1219.66 & 1307.34 & 831.05 & 1227.09 & 1440.15 & 841.94 & 804.92 \\
        abz9 & 1375.52 & 1184.63 & 1269.44 & 858.33 & 1193.77 & 1399.35 & 858.84 & 849.19 \\
        car1 & 8774.19 & 8551.98 & 9690.52 & 8423.45 & 8497.53 & 10135.55 & 6965.39 & 6990.89 \\
        car2 & 8690.52 & 8317.16 & 9270.20 & 7832.48 & 8321.70 & 9738.73 & 6742.86 & 6822.25 \\
\hline
    \end{tabular}
    \caption{Experimental results on Hurink's \cite{hurink1994tabu} \emph{edata}, with a time budget of 60 seconds.}
    \label{tab:edata_mean}
\end{table}

\clearpage

\begin{table}[!h]
    \tiny
    \centering
    \begin{tabular}{|c|c|c|c|c|c|c|c|c|}
         \hline
         Instance & NRPA & MCTS & NMCTS & HBNMCTS & BINRPA & RandGreedy & GNRPA & ABGNRPA \\
         \hline
        abz5 & 1891.25 & 1712.56 & 1956.19 & 1206.38 & 1716.08 & 2189.99 & 1113.81 & 1101.56 \\
        abz6 & 1461.58 & 1308.70 & 1494.52 & 970.53 & 1311.58 & 1708.67 & 914.59 & 913.16 \\
        abz7 & 1423.41 & 1217.55 & 1271.08 & 716.66 & 1217.02 & 1439.06 & 791.03 & 694.67 \\
        abz8 & 1489.42 & 1257.86 & 1318.66 & 741.81 & 1257.48 & 1487.41 & 814.02 & 725.08 \\
        abz9 & 1433.48 & 1225.86 & 1286.73 & 752.52 & 1225.78 & 1443.50 & 814.02 & 737.73 \\
        car1 & 7328.19 & 7212.58 & 8056.66 & 6281.98 & 7165.17 & 8186.89 & 5627.53 & 5688.83 \\
        car2 & 7919.41 & 7620.02 & 8366.36 & 6890.48 & 7574.45 & 8751.56 & 6426.31 & 6495.05 \\
\hline
    \end{tabular}
    \caption{Experimental results on Hurink's \cite{hurink1994tabu} \emph{rdata}, with a time budget of 60 seconds.}
    \label{tab:rdata_mean}
\end{table}

\clearpage

\begin{table}[!h]
    \tiny
    \centering
    \begin{tabular}{|c|c|c|c|c|c|c|c|c|}
         \hline
         Instance & NRPA & MCTS & NMCTS & HBNMCTS & BINRPA & RandGreedy & GNRPA & ABGNRPA\\
         \hline
        abz5 & 1753.33 & 1664.14 & 1790.33 & 942.42 & 1655.78 & 2093.46 & 1017.55 & 918.59 \\
        abz6 & 1374.70 & 1294.81 & 1406.89 & 821.77 & 1298.84 & 1657.11 & 851.88 & 791.39 \\
        abz7 & 1355.19 & 1188.02 & 1205.69 & 629.69 & 1192.33 & 1397.60 & 663.17 & 634.59 \\
        abz8 & 1434.38 & 1244.05 & 1252.42 & 630.97 & 1245.67 & 1471.03 & 738.08 & 634.88 \\
        abz9 & 1367.28 & 1195.95 & 1215.84 & 647.36 & 1189.09 & 1404.20 & 728.50 & 648.89 \\
        car1 & 7302.58 & 6861.69 & 7815.27 & 5845.20 & 6858.05 & 7871.34 & 5641.80 & 5682.13 \\
        car2 & 7618.09 & 7322.77 & 8152.55 & 6636.67 & 7349.72 & 8397.75 & 6324.17 & 6465.84 \\
\hline
    \end{tabular}
    \caption{Experimental results on Hurink's \cite{hurink1994tabu} \emph{vdata}, with a time budget of 60 seconds.}
    \label{tab:vdata_mean}
\end{table}

\begin{table}[!h]
    \begin{adjustbox}{width=\linewidth,center}
        \begin{tabular}{|c|c|c|c|c|c|c|c|c|}
            \hline
            Instance & NRPA & MCTS & NMCTS & HBNMCTS & BINRPA & RandGreedy & GNRPA & ABGNRPA\\
             \hline
            mk01 & 67.00 & 63.17 & 75.41 & 50.09 & 62.45 & 96.52 & 45.11 & 45.66 \\
            mk02 & 59.33 & 57.14 & 65.88 & 39.70 & 57.48 & 86.07 & 34.63 & 36.09 \\
            mk03 & 408.11 & 359.81 & 389.72 & 269.34 & 359.36 & 476.90 & 231.94 & 258.23 \\
            mk04 & 114.84 & 106.78 & 122.22 & 77.66 & 106.66 & 150.34 & 72.48 & 72.72 \\
            mk05 & 253.03 & 238.27 & 261.22 & 197.73 & 239.86 & 302.56 & 190.30 & 190.27 \\
            mk06 & 208.17 & 179.52 & 194.03 & 74.02 & 179.11 & 237.95 & 71.61 & 70.78 \\
            mk07 & 283.17 & 269.80 & 287.25 & 184.92 & 271.03 & 364.70 & 173.97 & 178.97 \\
            mk08 & 787.59 & 695.94 & 740.86 & 569.97 & 702.75 & 830.21 & 537.14 & 554.47 \\
            mk09 & 721.95 & 628.06 & 654.89 & 390.09 & 625.98 & 766.02 & 405.45 & 380.50 \\
            mk10 & 625.13 & 537.20 & 560.25 & 340.69 & 537.80 & 665.88 & 322.20 & 333.16 \\
            k1 & 17.02 & 13.05 & 21.89 & 12.45 & 13.20 & 34.18 & 11.00 & 11.00 \\
            k2 & 32.50 & 28.91 & 34.42 & 14.61 & 29.13 & 69.72 & 12.00 & 11.97 \\
            k3 & 26.73 & 24.53 & 28.64 & 8.11 & 24.61 & 48.41 & 7.19 & 7.03 \\
            k4 & 47.72 & 45.98 & 51.16 & 14.27 & 46.66 & 65.07 & 12.59 & 13.47 \\
            \hline
        \end{tabular}
    \end{adjustbox}
    \caption{Experimental results on Kacem's \cite{kacem2002pareto} and Brandimarte's \cite{brandimarte1993routing} instances: average makespan on 100 iterations of each algorithm with a time budget of 60 seconds for each iteration.}
    \label{tab:kacem_brandimarte_mean}
\end{table}

\clearpage

\subsection{Standard deviation of the makespan}

\begin{table}[]
    \centering
    \begin{tabular}{|c|c|c|c|c|c|c|c|}
        \hline
        NRPA & MCTS & NMCTS & HBNMCTS & BINRPA & RandGreedy & GNRPA & ABGNRPA \\
        \hline
        6.89 & 4.08 & 5.94 & 3.30 & 4.17 & 7.93 & 2.09 & \textbf{1.61} \\
        \hline
    \end{tabular}
    \caption{Average ranking of tested algorithms on the standard deviation of the makespan obtained on 100 iterations with a budget of 60 seconds on Hurink's, Kacem's, and Brandimarte's instances}
    \label{tab:std_makespan_ranking}
\end{table}

\begin{table}[!h]
    \tiny
    \centering
    \begin{tabular}{|c|c|c|c|c|c|c|c|c|}
        \hline

        Instance & NRPA & MCTS & nmcts & HBNMCTS & BINRPA & RandGreedy & GNRPA & ABGNRPA \\
        \hline
        abz5 & 139.44 & 47.50 & 100.15 & 55.17 & 50.87 & 195.08 & 18.90 & 14.11 \\
        abz6 & 133.32 & 37.59 & 80.93 & 32.38 & 31.34 & 152.45 & 96.49 & 9.73 \\
        abz7 & 76.62 & 29.51 & 44.19 & 18.31 & 27.66 & 99.57 & 12.78 & 13.37 \\
        abz8 & 100.14 & 32.20 & 51.68 & 20.64 & 24.35 & 100.25 & 18.14 & 11.33 \\
        abz9 & 78.65 & 30.18 & 38.45 & 15.09 & 27.02 & 96.85 & 17.38 & 12.97 \\
        car1 & 720.27 & 266.85 & 870.88 & 417.85 & 313.23 & 895.63 & 13.29 & 98.80 \\
        car2 & 707.89 & 196.23 & 573.26 & 360.23 & 223.54 & 790.31 & 14.95 & 88.50 \\
\hline
    \end{tabular}
    \caption{Experimental results on Hurink's \cite{hurink1994tabu} \emph{edata}, with a time budget of 60 seconds.}
    \label{tab:edata_std}
\end{table}

\clearpage

\begin{table}[!h]
    \tiny
    \centering
    \begin{tabular}{|c|c|c|c|c|c|c|c|c|}
         \hline
         Instance & NRPA & MCTS & NMCTS & HBNMCTS & BINRPA & RandGreedy & GNRPA & ABGNRPA \\
         \hline

        abz5 & 173.31 & 57.75 & 86.47 & 39.04 & 52.36 & 210.20 & 21.61 & 20.61 \\
        abz6 & 104.32 & 46.48 & 86.52 & 24.82 & 43.94 & 172.62 & 78.08 & 14.03 \\
        abz7 & 90.91 & 27.78 & 45.56 & 17.17 & 33.70 & 106.26 & 15.45 & 15.95 \\
        abz8 & 96.27 & 33.10 & 45.72 & 15.35 & 32.23 & 110.60 & 13.09 & 12.16 \\
        abz9 & 89.39 & 31.28 & 39.51 & 11.43 & 29.65 & 103.98 & 19.11 & 10.19 \\
        car1 & 527.04 & 202.97 & 498.77 & 231.49 & 223.59 & 706.92 & 10.39 & 66.65 \\
        car2 & 578.36 & 175.20 & 513.97 & 224.37 & 206.92 & 720.62 & 14.16 & 76.32 \\
\hline
    \end{tabular}
    \caption{Experimental results on Hurink's \cite{hurink1994tabu} \emph{rdata}, with a time budget of 60 seconds.}
    \label{tab:rdata_std}
\end{table}

\clearpage

\begin{table}[!h]
    \tiny
    \centering
    \begin{tabular}{|c|c|c|c|c|c|c|c|c|}
         \hline
         Instance & NRPA & MCTS & NMCTS & HBNMCTS & BINRPA & RandGreedy & GNRPA & ABGNRPA\\
         \hline

        abz5 & 130.86 & 48.01 & 89.13 & 15.92 & 58.34 & 205.11 & 19.69 & 10.03 \\
        abz6 & 98.09 & 39.59 & 71.95 & 16.33 & 43.57 & 
        163.30 & 83.48 & 11.05 \\
        abz7 & 71.12 & 33.91 & 32.27 & 10.18 & 33.90 &
        104.98 & 13.83 & 12.94 \\
        abz8 & 75.24 & 33.89 & 38.38 & 8.05 & 37.82 & 115.79 & 10.00 & 9.21  \\
        abz9 & 71.88 & 30.09 & 32.20 & 9.47 & 37.76 &
        105.15 & 17.26 & 9.72 \\
        car1 & 677.85 & 200.17 & 470.40 & 152.02 & 195.85 & 733.74 & 12.60 & 90.75 \\
        car2 & 555.98 & 193.94 & 401.64 & 195.55 & 206.95 & 716.50 & 12.38 & 73.86 \\
\hline
    \end{tabular}
    \caption{Experimental results on Hurink's \cite{hurink1994tabu} \emph{vdata}, with a time budget of 60 seconds.}
    \label{tab:vdata_std}
\end{table}

\begin{table}[!h]
\begin{adjustbox}{width=\linewidth,center}
    \begin{tabular}{|c|c|c|c|c|c|c|c|c|}
        \hline
        Instance & NRPA & MCTS & NMCTS & HBNMCTS & BINRPA & RandGreedy & GNRPA & ABGNRPA\\
         \hline
        mk01 & 5.87 & 2.25 & 6.91 & 2.64 & 2.28 & 11.98 & 9.57 & 0.71 \\
        mk02 & 4.41 & 1.78 & 3.64 & 2.52 & 2.09 & 10.90 & 4.46 & 0.96 \\
        mk03 & 31.34 & 10.09 & 15.92 & 10.96 & 11.99 & 45.92 & 4.65 & 6.01 \\
        mk04 & 10.60 & 3.59 & 7.12 & 2.43 & 3.76 & 16.60 & 0.95 & 0.94 \\
        mk05 & 17.34 & 5.42 & 8.87 & 4.23 & 5.79 & 22.93 & 1.02 & 2.32 \\
        mk06 & 17.20 & 6.48 & 9.20 & 1.63 & 6.13 & 22.33 & 16.84 & 1.14 \\
        mk07 & 22.27 & 7.40 & 14.19 & 5.84 & 9.60 & 37.19 & 5.84 & 2.89 \\
        mk08 & 54.96 & 16.65 & 26.19 & 9.65 & 15.27 & 54.40 & 0.85 & 5.26 \\
        mk09 & 47.75 & 15.66 & 27.00 & 10.79 & 18.97 & 57.87 & 6.18 & 7.31 \\
        mk10 & 42.47 & 14.64 & 18.92 & 12.56 & 12.90 & 52.33 & 6.27 & 8.56 \\
        k1 & 2.56 & 0.80 & 5.60 & 2.40 & 0.71 & 18.52 & 0.00 & 0.00 \\
        k2 & 4.10 & 1.70 & 3.32 & 1.02 & 1.49 & 32.59 & 0.00 & 0.25 \\
        k3 & 3.63 & 1.20 & 2.48 & 0.36 & 1.54 & 10.55 & 0.39 & 0.17 \\
        k4 & 9.09 & 2.36 & 3.12 & 0.48 & 2.55 & 32.13 & 0.49 & 0.50 \\
        \hline
    \end{tabular}
    \end{adjustbox}
    \caption{Experimental results on Kacem's \cite{kacem2002pareto} and Brandimarte's \cite{brandimarte1993routing} instancesmakespan standard deviation on 100 iterations of each algorithm with a time budget of 60 seconds for each iteration.}
    \label{tab:kacem_brandimarte_std}
\end{table}

\clearpage

\subsection{Bias deviation of ABGNRPA}
\label{bias_deviation_analysis}

\begin{table}[!h]
    \centering
    %
    \caption{Bias deviation between the start and the end of an ABGNRPA run in percentages for Hurink's instances \cite{hurink1994tabu}.}
    \label{tab:hurink_bias_difference}
\end{table}

\begin{table}[]
    \centering
    \begin{tabular}{|c|c|}
        \hline
        Instance & Deviation \\
        \hline
        k1 & 23.59\% \\
        \hline
        k2 & 28.06\% \\
        \hline
        k3 & 18.43\% \\
        \hline
        k4 & 13.38\% \\
        \hline
        mk01 & 22.79\% \\
        \hline
        mk02 & 18.71\% \\
        \hline
        mk03 & 13.52\% \\
        \hline
        mk04 & 21.80\% \\
        \hline
        mk05 & 21.05\% \\
        \hline
        mk06 & 10.19\% \\
        \hline
        mk07 & 30.07\% \\
        \hline
        mk08 & 13.09\% \\
        \hline
        mk09 & 10.87\% \\
        \hline
        mk10 & 11.01\% \\
        \hline
    \end{tabular}
    \caption{Bias deviation between the start and the end of an ABGNRPA run in percentages for Kacem \cite{kacem2002pareto} and Brandimarte \cite{brandimarte1993routing} instances.}
    \label{tab:kacem_brandimarte_bias_deviation}
\end{table}

\end{document}